\documentclass[letterpaper, 10 pt, conference]{ieeeconf}  

\IEEEoverridecommandlockouts                              

\usepackage[utf8]{inputenc}
\usepackage[T1]{fontenc}

\usepackage{amsmath} 
\usepackage{graphicx} 
\usepackage{booktabs, siunitx, etoolbox}
\usepackage{multirow}
\usepackage{cite}
\usepackage{xcolor}
\usepackage{hyperref}
\usepackage{bm}
\usepackage{mathtools}
\usepackage[font=normalsize]{caption}

\title{\LARGE \bf Semantically Grounded Object Matching \\ for Robust Robotic Scene Rearrangement}

\author{Walter Goodwin$^{1}$, Sagar Vaze$^{2}$, Ioannis Havoutis$^{1}$ and Ingmar Posner$^{1}$
\thanks{$^{1}$Oxford Robotics Institute, $^{2}$Visual Geometry Group, University of Oxford, $^{1}$23/$^{2}$25 Banbury Rd, Oxford OX2 6NN}
\thanks{\tt\small firstname@robots.ox.ac.uk}%
}

\begin{document}
\renewcommand{\baselinestretch}{0.9}
\maketitle
\thispagestyle{empty}
\pagestyle{empty}


\begin{abstract}
Object rearrangement has recently emerged as a key competency in robot manipulation, with practical solutions generally involving object detection, recognition, grasping and high-level planning. Goal-images describing a desired scene configuration are a promising and increasingly used mode of instruction. A key outstanding challenge is the accurate inference of matches between objects in front of a robot, and those seen in a provided goal image, where recent works have struggled in the absence of object-specific training data.
In this work, we explore the deterioration of existing methods' ability to infer matches between objects as the visual shift between observed and goal scenes increases. We find that a fundamental limitation of the current setting is that source and target images must contain the same \textit{instance} of every object, which restricts practical deployment. We present a novel approach to object matching that uses a large pre-trained vision-language model to match objects in a cross-instance setting by leveraging semantics together with visual features as a more robust, and much more general, measure of similarity. We demonstrate that this provides considerably improved matching performance in cross-instance settings, and can be used to guide multi-object rearrangement with a robot manipulator from an image that shares no object \textit{instances} with the robot's scene. 
Our code is available at \url{https://github.com/applied-ai-lab/object_matching}.
\end{abstract}

\section{INTRODUCTION} \label{sec:intro}
In recent years, there have been a number of successes in applying deep learning to enable manipulation skills on robots which operate directly from images. Alongside this, the `goal image' has emerged as a convenient way to specify an instruction for a robotic task, ranging from guiding visual servoing towards a single object \cite{Sadeghi2018} to multi-object rearrangement settings \cite{Batra2020, Qureshi2021_NeRP, Groth2021}. Goal images enable specification of the goal in the same modality as the system input for a robot with vision sensors, and are a practical way to express desired spatial outcomes for objects in a scene in many settings. Consider a household task, such as laying a dining table, or an industrial task, such as kitting products into a canonical set --- in both settings it can be more convenient to simply image the desired goal state rather than exhaustively define all degrees of freedom.

Furthermore, object rearrangement problems have attracted interest recently as a challenging, and quite general, robotic manipulation problem, with the goal image motivated as a key component ~\cite{Batra2020, Szot2021, Weihs2021, Liu2021, Qureshi2021_NeRP, Danielczuk2021}. There has been some success recently in the development of systems that can achieve rearrangement tasks when provided with a goal image \cite{Qureshi2021_NeRP, Labbe2020}, generally through the integration of pre-trained object segmentation, grasping primitives or grasp prediction networks, and subsequent high-level planning at the object level. However, progress in learning both grasp prediction networks \cite{Morrison2020, Jiang2021} and segmentation networks \cite{XiangUCN2020} that generalise to unseen objects, mean that it is increasingly possible to both localise and grasp objects (two important stages in successful rearrangement ~\cite{Qureshi2021_NeRP, Danielczuk2021}). 

Thus, a key outstanding challenge in specifying outcomes with images is that of robustly \textit{matching} objects in a goal image to those in the current scene, following object detection. Successful matching is critical to successful interpretation of a goal image, and poor matching has emerged as the principal performance bottleneck in recent works on tabletop rearrangement \cite{Qureshi2021_NeRP, Zhang2021ACRV}. Currently, a wide range of matching methods are used, from hand crafted features such as colour histograms \cite{Swain1991} to object-specific visual features learnt unsupervised \cite{WuAPEX2021}, and deep feature extractors trained on large scale computer vision datasets \cite{Labbe2020, Qureshi2021_NeRP}. 

This work conducts a controlled analysis and comparison of approaches to matching under various visual shifts between the current (source) and goal (target) images, which we find to be lacking in the existing literature. We control for factors such as pose and background mismatch and report results on a number of baselines. Furthermore, we identify a fundamental limitation of the existing matching approaches, finding that matching is currently only demonstrated when source and target images contain the same \textit{instance}, i.e two views of exactly the same object. This can be restrictive in practical settings. Consider again the table laying problem. It is unreasonable to expect the goal image to contain the precise crockery instances in the current scene. Rather, the goal image will specify a \textit{semantically} consistent target, specifying where a mug or plate should go, regardless of their precise appearance. We thus further examine the matching problem in a cross-instance setting, in which source and target images contain different instances of the same \textit{semantic} object, which may be visually distinct. In this setting, we find that existing matching approaches, which rely solely on visual descriptors, degrade significantly in performance.

Next, we propose a novel solution to the cross-instance matching problem via the re-purposing of the recently released CLIP model \cite{Radford2021}. In contrast to existing matching approaches, which rely solely on visual features, the proposed model facilitates the introduction of semantic information to the matching problem, leading to a substantial boost in our proposed cross-instance matching performance. Finally, we deploy our method on a real robot system, finding that our proposed semantic matching protocol is important for cross-instance matching in the real world.

\begin{figure*}[ht!]
\includegraphics[width=\textwidth]{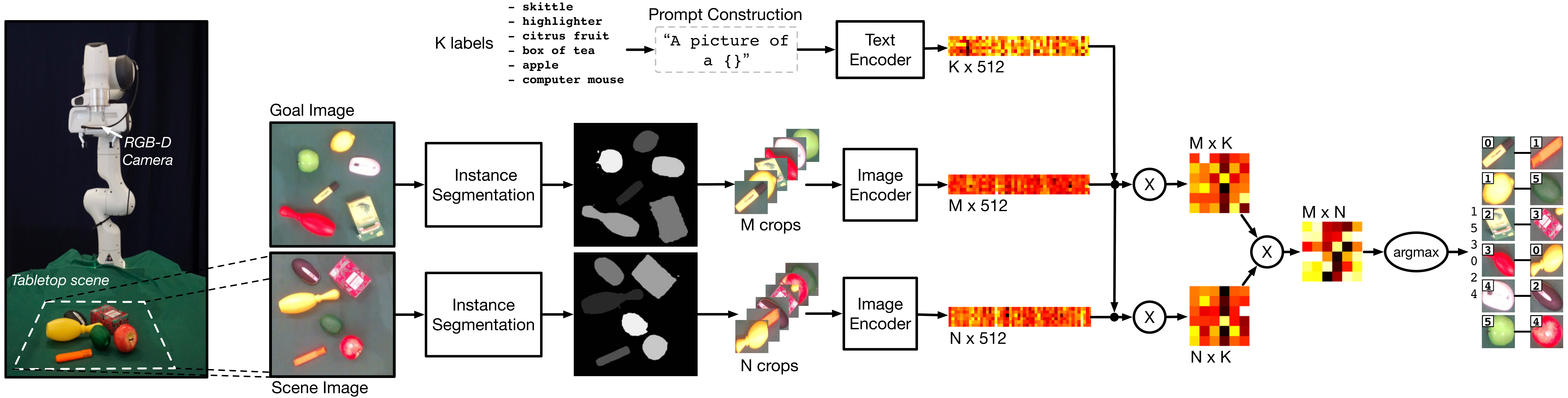}
  \vspace{-1.7em}
  \caption{Visual-Semantic Matching: For both current scene and goal images, we use an instance segmentation network to extract crops of all present objects. We then compute similarities between the crops' CLIP visual features and the semantic features of a set of $K$ text prompts. A final $M \times N$ similarity matrix between the current and goal crops can then be computed based on the crops' similarity to each of the semantic features. The `X' operator is $X(\mathbf{A}, \mathbf{B}) \coloneqq \mathbf{AB^T}$.}
  \vspace{-1.75em}
  \label{fig:semfeat}
\end{figure*}
\section{RELATED WORK} \label{sec:related_work}
A number of recent works, challenges and benchmarks have considered robotic scene rearrangement with visual observations, a challenging problem for robotics and embodied artificial intelligence \cite{Batra2020}. Simulated benchmarks have been proposed for both mobile \cite{Szot2021, Weihs2021} and tabletop robotic manipulation \cite{James2020RLBench, Liu2021}. A common theme is the use of visual instructions in the form goal images, as a means of communicating a target scene configuration. Recent attempts at tabletop rearrangement on \textit{real} robot platforms take a modular approach, where object detection is executed in the current (source) scene and the goal (target) image. Objects are then matched across the scenes, and planning over pick-and-place actions brings about the desired object displacements \cite{Qureshi2021_NeRP, Labbe2020, Zhang2021ACRV, Danielczuk2021}. Existing works have tackled particular parts of this pipeline, with work on improved collision checking \cite{Danielczuk2021}, high-level planning \cite{Qureshi2021_NeRP, Labbe2020, Song2020}, and vision-based RL approaches \cite{Yuan2018, Groth2021}.
Several of these works note failures in successful matching \cite{Qureshi2021_NeRP, Labbe2020} due to the challenges of accurately comparing objects from only single views. It is notable that, with the exception of some invariance to viewpoint shift \cite{Labbe2020},  goal-conditioned works are restricted to goal images generated in the exact setting, and with the exact object instances, that will be encountered by the robot at deployment. Matching approaches handle only \textit{instance-level} correspondence, and cannot infer \textit{semantic} matches between scenes. We review existing matching approaches, and attempts to exploit object semantics, in robotics.

\subsection{Object matching in robotics}
Hand-crafted visual descriptors such as colour histograms were originally proposed for this task \cite{Swain1991} and remain widely used today \cite{iRobotCNC2020}. More recently, features extracted from the backbone of a CNN classifier have been applied to tackle the matching problem \cite{Labbe2020, Qureshi2021_NeRP}. These features have been trained on large scale datasets such as ImageNet \cite{Deng2009} and demonstrate invariance to nuisance factors such as lighting and 2D rotation. However, outside of training classes, these features have not been trained to be invariant to \textit{instance-shift}, and hence degrade in matching performance when the source and target images contain different instances of semantically identical objects.
Object matching is related to the mature field of template matching, but where only one single view of the object exists as the `template'. Solutions to single-view instance recognition have been proposed. \cite{Held2016} fine-tune a CNN pre-trained on ImageNet on multiple views of many 3D objects to learn pose-invariant representations which is finetuned on a single image of a new object for pose-invariant object detection. They show that this performs far better than a large range of pre-existing template matching techniques for a single template image. Other deep learning methods reduce the number of views required \cite{Mercier2021, Mercier2017, Ammirato2018}. However, while these methods are often able to build models robust to \textit{pose} shift of objects, they require that the same \textit{instance} of the target object is being sought. \cite{Zeng2018} trained a cross-domain image matching technique to enable products viewed by the robot to be matched to a database of Amazon product photos, but this \textit{instance}-matching method requires a known, \textit{systematic} shift between source and target. In contrast, in this work, through leveraging visual-semantic grounding, we handle a much more general case, in which arbitrarily different instances of objects with the same underlying semantic description can be successfully paired.

\subsection{Visual-semantic object picking}
There have been numerous attempts to enable language-guided robotic manipulation by grounding language instructions to visual observations. While a different mode of instruction to the goal image, these works are related in their use of semantics in aiding robotic disambiguation of the visual world, and in their  use of vision-language models. Work concurrent with this \textit{conditions} imitation learning on CLIP embeddings of text instructions to improve generalisation \cite{Shridhar2021}. Several recent works present systems to guide a robot to pick a particular object from a scene with language prompts viewed as \textit{referring expressions} \cite{Zhang2021INVIGORATE, Mees2021, Hatori2018}. Referring expression comprehension is an area of computer vision research that seeks to ground an unstructured text prompt from a human that \textit{refers} to an item visually present in an image, and locate the item on this basis. When brought to robotic systems, standard models \cite{Yu2018MAttNet} trained on a dataset with a limited number of classes tend to be used \cite{Zhang2021INVIGORATE, Mees2021, Hatori2018}. While this work also proposes language as a mechanism for resolving visual ambiguities, we do not rely on pre-training and are able to handle arbitrary object classes. 

\section{Visual Object Matching} \label{sec:methods}

In this section we describe the core matching models we compare in this paper. We first describe the proposed visual-semantic models based on CLIP \cite{Radford2021}, before summarising the baseline matching models against which we compare. 

\textbf{Preliminary notation} A common operation when comparing objects is, given a set of inputs, $\mathcal{T}$, and a feature extraction function, $f(\cdot)$, to build a set of normalised feature vectors. Here $\mathcal{T}$ could refer to a set of image crops or text prompts. Formally, we define the operation $\mathcal{F}$, as:

\begin{equation}
    \mathcal{F}(f, \mathcal{T}) = \Big\{\frac{f(t)}{|f(t)|} \quad  \forall t \in \mathcal{T}\Big\}
\end{equation}

\subsection{Visual-Semantic Models} \label{sec:exp-classification}

In this work, we propose \textit{semantic matching} through the recently released CLIP model \cite{Radford2021}. CLIP consists of a pair of neural network embeddings which jointly map text-image pairs into a common feature space. The model is trained on web-scale data and is thus capable of interpreting a wide range of semantic objects. In this way, by finding similarities between a set of text (semantic) prompts and a given image, one can identify the most likely category of an object within the set in a `zero-shot' fashion.   

\subsubsection{Object Matching} \label{sec:method-semantic}

The proposed visual-semantic object matching process is shown in Figure \ref{fig:semfeat}. Leveraging unseen object instance segmentation \cite{XiangUCN2020}, object crops are taken, and all $M$ objects in the source and the $N$ objects in the target are passed through the image encoder. The set of $K$ possible object categories are passed through the text encoder. This allows us to construct two classification matrices, $C_s$ and $C_t$, which describe the model's confidence that each object belongs to each of the $K$ categories. Next, based on these confidences for each object, we compute a similarity between each object in the source and the target. 

Formally, consider $\Phi_{v}$ and $\Phi_{s}$ as the deep image and text CLIP encoders respectively. Further consider $\mathcal{X}$ is the set of cropped object patches in either the source, $\mathcal{X}_s$, or target, $\mathcal{X}_t$, and $\mathcal{Y}_K$ as the set of semantic text prompts. A set of normalised visual features is constructed for the source and target crops as $v_s = \mathcal{F}(\Phi_{v}, \mathcal{X}_s)$ and $v_t = \mathcal{F}(\Phi_{v}, \mathcal{X}_t)$, and semantic features are extracted as $s = \mathcal{F}(\Phi_{s}, \mathcal{Y}_K)$.

The classification matrix for either source or target, $C$, is then constructed as $C_{ik} = \langle\, v_i , s_k \rangle$ where $\langle\cdot,\cdot\rangle$ represents an inner-product. The similarity matrix $S$ between source and target is computed as $S = C_{t}C_{s}^{T}$. The final step is performing assignment based on the similarity matrix, which can either be performed with the \texttt{argmax()} operation or through minimum weight matching with the Hungarian algorithm. We experiment with two variants of the CLIP-Semantic model: first we pass all $K$ semantic labels for matching (\textbf{CLIP-SemFeat-K}); we also pass only the semantic labels which we know to be present in the target image (\textbf{CLIP-SemFeat-N}). These two settings both correspond to practical scenarios. In many tasks, there might not be prior knowledge of exactly what will be in the scene, but we know that the relevant objects can be described by a subset of $K$ labels. For instance, a robot laying a table might enumerate the names of all items of tableware, even if only a fork and spoon are present. On the other hand, in a warehousing setting, the names of the objects present may be known, giving a set of $N$ labels. Intuitively, if $K>N$, \textbf{CLIP-SemFeat-K} considers some labels irrelevant to the scene. 

\subsubsection{Prompt Engineering}

We seek to optimise CLIP to ground the textual descriptions of the objects to their corresponding objects across the considered datasets (Section \ref{sec:datasets}), to ensure that the model can provide meaningful proximity measures between crops of objects across the images, and their text descriptions, through the shared embedding space. To this end, we engineer the textual prompts used in the semantic embedding function, choosing them such that a small selection of crops show a small distance between the semantic and visual embedding.
For each object class, we do this by taking a small number of reference crops, and typing around 5 short descriptions of the object. Using CLIP, we compute the cosine similarity between the visual embedding of the crops, and text embeddings of these descriptions (including the original description). We take the descriptions with highest similarity as the improved set of semantic prompts, with the entire process taking a couple of minutes for a single user.

\subsection{Baselines}

In our experiments, we compare to a number of baseline matching approaches used in recent rearrangement works. All methods rely on extracting purely visual descriptors of crops based on the pixel values, before constructing a similarity matrix between the $N$ source and $M$ target crops. 

Formally, the visual features are extracted from the crops as $v = \mathcal{F}(f, \mathcal{X})$ for a given feature extraction function $f(\cdot)$. The similarity matrix is computed solely based on the visual descriptors as $S_{mn} = \langle\, v_m , v_n \rangle$.

\textbf{Colour Histograms} The winning submission from 59 teams to the OCRTOC Tabletop Rearrangement Challenge 2020 used the cosine similarity between colour histograms for object recognition, and deployed nearest neighbour voting against a dataset of multiple images of each potential object to determine the identity of an object in a scene \cite{iRobotCNC2020}. While the goal-image setting provides only one reference crop for each object to be matched, we consider cosine distance on colour histograms as a first baseline. Here, $f(\cdot)$ involves concatenating histograms for both the hue and saturation values across all pixels in an image. Hue and saturation are projections of RGB pixel values into a frame more in line with visual cues of interest to human observers \cite{Swain1991}.

\textbf{AlexNet-S} In \cite{Labbe2020}, which achieves multi-object rearrangement with goal-image matching for up to 12 cubes on a tabletop, the authors use distances between the output of the conv3 layer of AlexNet pre-trained on ImageNet. The objects considered, though, are all cubes with distinct colours. Here, $f(\cdot)$ extracts the flattened spatial features of the conv3 layer, which retain some of the spatial structure of the input crops.

\textbf{ResNet50-(S/G)} Most recently, ResNet50 features have been used in \cite{Qureshi2021_NeRP}, where they were used for finding object matches between scenes of 2 to 5 objects. In this case, we experiment with two settings, with spatial features, \textbf{ResNet50-S} (before the final max-pool layer), and with global features \textbf{ResNet50-G} (immediately after the final max-pool layer).

\textbf{CLIPVisual} To allow us to isolate the effect of the \textit{semantic} information encoded in the full CLIP model, from the effect of the CLIP model's visual feature extractor alone, we also embed the crops through only the visual CLIP backbone, $f = \Phi_{v}$. We use these features identically to the other purely visual descriptors.

\section{DATASETS} \label{sec:datasets}
\begin{figure}[tbp]
\centering
  \includegraphics[width=0.98\columnwidth]{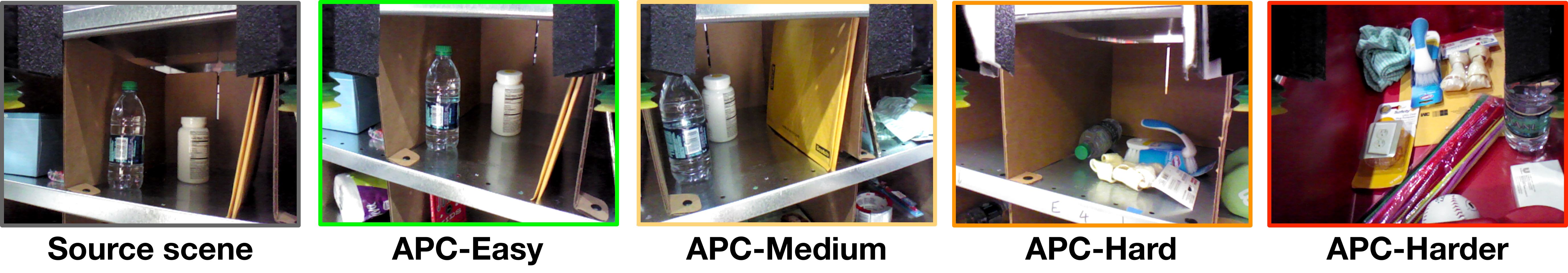}
  \caption{Examples of the 4 same-instance matching settings we construct from the APC dataset.}
  \label{fig:apc-data}
\end{figure}
Our aim is to investigate matching performance in two distinct settings: \emph{same-instance matching} and \emph{cross-instance matching}. We first leverage the APC dataset \cite{Zeng2017APCPose}, which allows investigation of the instance matching setting while controlling for degrees of visual shifts in pose and background. We then use the LVIS dataset \cite{Gupta2019} to look at cross-instance matching, as it contains multiple instances of the same semantic class in different settings.
\begin{table}[tbp]
\vskip 0.15in
\begin{center}
\begin{small}
\begin{sc}
\sisetup{detect-weight,mode=text}
\renewrobustcmd{\bfseries}{\fontseries{b}\selectfont}
\renewrobustcmd{\boldmath}{}
\newrobustcmd{\B}{\bfseries}
\begin{tabular}{l|
    S[table-format = 2.1]
    S[table-format = 2.1]
}
\toprule
Model & \text{Top-1} & \text{Top-5} \\
\hline
Random Guess & 2.5641 & 12.8205 \\
\textbf{CLIP} & 30.7036 & 54.5788 \\
\textbf{CLIP+} & \B 38.0105 & \B 65.2276 \\
\textbf{CLIP++} & 34.9579 & 61.4962 \\
\bottomrule
\end{tabular}
\end{sc}
\end{small}
\end{center}
\vskip 0.1in
\caption{Zero-shot classification performance of the vision-language model (CLIP) across the APC dataset. \textbf{CLIP} uses the exact same wording as the original product name from APC \cite{Zeng2017APCPose}, formatted as "A picture of a \{...\}". \textbf{CLIP+} uses the same formatting but with `better' labels chosen as described in \ref{sec:prompteng}. \textbf{CLIP++} additionally ensembles over all labels and additional prompt formats, as described in \ref{sec:prompteng}.}
\label{table:classification}
\end{table}

\subsection{Amazon Picking Challenge (APC) 2016 dataset} \label{sec:apc-exp}
We use the dataset collected by the MIT team for their entry to the Amazon Picking Challenge (APC) 2016 \cite{Zeng2017APCPose}. This comes with accurate predicted instance segmentation masks for the 39 objects considered in APC 2016, and contains 7,281 images from 452 distinct scenes. Scenes contain between one and twelve objects arranged in two different settings: a shelf, pictured from the side and a plastic tote box, pictured from above. Each scene type was recorded in two different locations, with different lighting conditions. Each shelf scene is imaged from 15 different views, and each tote scene from 18 views. The same 39 object instances occur throughout, though between scenes they vary in pose, occlusion relationships with other objects, background and lighting conditions. From these conditions we are able to form four different object matching settings, which we hypothesise - and empirically confirm - pose progressively more challenging conditions for matching. \\
\noindent{\bf APC-Easy}: we set up matching problems where both source and target image are drawn from exactly the same scene, but from different views. In this condition, we consider only pairs over views that are close. Objects will in general retain substantial visual similarity, but will be viewed from different angles, and occlusion conditions may change slightly.\\
\noindent{\bf APC-Medium}: as with \textbf{APC-Easy}, except we consider source/target pairs for matching that are maximally dissimilar i.e. viewed from diagonally opposite corners. While still matching within-scene, relative object poses are substantially shifted, and occlusion conditions will vary. \\
\noindent{\bf APC-Hard}: we formulate source-target pairs from \textit{different} scenes but of the same setting (e.g. shelf). All scene pairs in which there is at least one valid match to be made are matched. Object poses are different between scenes, and some objects will not have any valid match - we do not count these towards the reported accuracy.\\
\noindent{\bf APC-Hardest}: as with \textbf{APC-Hard}, but we sample source and target scenes from opposite settings e.g. shelf vs tote.

We remove any trivial examples from these partitions, where both source and target scene contain just one object. 
\begin{table*}[t]
\centering
\vskip 0.15in
\begin{center}
\begin{small}
\begin{sc}
\sisetup{detect-weight,mode=text}
\renewrobustcmd{\bfseries}{\fontseries{b}\selectfont}
\renewrobustcmd{\boldmath}{}
\newrobustcmd{\B}{\bfseries}
\begin{tabular}{l|
    S[table-format = 2.1]
    S[table-format = 2.1]
    S[table-format = 2.1]
    S[table-format = 2.1]|
    S[table-format = 2.1]
    S[table-format = 2.1]}
\toprule
\multirow{2}{*}{model} & \multicolumn{4}{c|}{Same-Instance (APC) \cite{Zeng2017APCPose}} & \multicolumn{2}{c}{Cross-Instance (LVIS) \cite{Gupta2019}}  \\
& \multicolumn{1}{c}{Easy {[}\%{]}} & \multicolumn{1}{c}{Medium {[}\%{]}} & \multicolumn{1}{c}{Hard {[}\%{]}} & \multicolumn{1}{c|}{Hardest {[}\%{]}} & \multicolumn{1}{c}{8-way {[}\%{]}} & \multicolumn{1}{c}{20-way {[}\%{]}} \\
\hline
RandomGuess & 32.2926 & 33.0626 & 25.3606 & 26.2815 & 12.395 & 4.9613 \\
ColourHist & 89.3702 & 57.6965 & 48.0000 & 42.1658 & 20.2881 & 9.8227 \\
AlexNet-S & 95.8428 & 60.6256 & 46.3061 & 37.7249 & 30.4987 & 17.6051 \\
ResNet50-S & \B 97.8838 & \B 68.7638 & 52.5669 & 42.6634 & 34.6494 & 21.0456 \\
ResNet50-G & 96.9119 & 67.6561 & \B 61.2773 & \B 54.9769 & 50.5569 & 35.4535 \\
CLIPvisual & 91.2411 & 60.7234 & 50.9236 & 46.095 & 40.3481 & 27.3657 \\
\hline
CLIPsemfeat-N & 81.8377  & 59.9634  & 52.7236  & 51.8012  & \B 58.775  & \B 40.1406 \\ 
CLIPsemfeat-K & 90.6487  & 63.7355  & 54.1719  & 52.0877  & 52.9075  & 37.7519  \\ 

\bottomrule
\end{tabular}
\end{sc}
\end{small}
\end{center}
\caption{
Matching success accuracy for baselines and visual-semantic models. The settings for same instance and different instance matching are as described in Sections \ref{sec:apc-exp}, \ref{sec:lvis-exp}. Best performing approaches are bold-faced. For increasingly difficult matching settings, the performance of approaches based on purely visual features deteriorates rapidly, with the \textbf{CLIP-SemFeat} approaches outperforming all baselines in cross-instance settings. 
}
\label{table:exp-main}
\end{table*}

\subsection{Large Vocabulary Instance Segmentation (LVIS) dataset} \label{sec:lvis-exp}

A key limitation of the current matching tasks demonstrated in the robotics literature is that they only consider same-instance correspondence. This is evident when looking at recent work \cite{Batra2020, Qureshi2021_NeRP, Groth2021} and instance-specific challenges and datasets such as APC \cite{Zeng2017APCPose}, and is likely due to the open nature of the cross-instance matching problem. However, achieving this goal would enable carrying out robotic manipulation tasks that leverage a goal image without the restrictive constraint that this goal (target) image needs to contain visually very similar objects to - i.e. the same object instance as -  the current (source) image. We present our lab-based experiments in Section \ref{sec:robot-exp}, but for a more comprehensive, in-the-wild, and unbiased assessment of visual-semantic matching performance on cross-instance settings, we use the Large Vocabulary Instance Segmentation (LVIS) dataset \cite{Gupta2019}, a densely annotated object recognition and segmentation dataset, with a training set of 1.27M annotations over 1203 object classes, across 100,170 images. Annotations consist of segmentation masks and class id. To formulate a large set of matching problems relevant to robotics applications, we select a subset of 40 objects that could be feasibly grasped by a tabletop manipulator, from the top 200 most-occurring classes. All annotations that have a pixel area of less than $32^2$ are disregarded, and and only the first instance of a given object class from each image is kept. This leaves around 36000 annotated instances, with a mean of 900 per class. Unlike in the APC dataset, the open-world settings and high number of possible classes in LVIS means most pairs of scenes have either zero or a small intersection of classes present. These scene pairs would present trivial matching settings, and so we formulate matching problems synthetically. For an N-way matching problem, we sample N labels from the set of 40 classes, and then sample a pair of different object crops $C^{S}_{i}, C^{T}_{i}$ for each label $i \in N$. We then seek to match the set of source crops $\{C^{S}_{i}\}^{N}_{1}$ against the set of target crops $\{C^{T}_{i}\}^{N}_{1}$. We crop based on ground-truth masks, as we focus on matching and not detection.

\begin{table}[htb]
\centering
\vskip 0.15in
\begin{center}
\begin{small}
\begin{sc}
\sisetup{detect-weight,mode=text}
\renewrobustcmd{\bfseries}{\fontseries{b}\selectfont}
\renewrobustcmd{\boldmath}{}
\newrobustcmd{\B}{\bfseries}
\begin{tabular}{l|
    S[table-format = 2.1]
    S[table-format = 2.1]
}
\toprule
\multicolumn{1}{c|}{model} & \multicolumn{1}{c}{Same- {[}\%{]}} & \multicolumn{1}{c}{Cross-instance {[}\%{]}} \\
\hline
RandomGuess & 15.6962 & 15.3226  \\
ColourHist & 69.6203 & 13.7097  \\
AlexNet-S & 58.9873 & 19.3548  \\
ResNet50-S & 57.7215 & 37.9032 \\
ResNet50-G & \B 76.5 & 55.6 \\
CLIPvisual & 72.9 & 49.8  \\
\hline
CLIPsemfeat-N & 70.1 & 74.2  \\
CLIPdiscrete-N & 59.0 & \B 77.4  \\
\bottomrule
\end{tabular}
\end{sc}
\end{small}
\end{center}
\caption{
Source to target image object accuracy for baselines and visual-semantic models across same-instance and different-instance robotic rearrangement scenes.
}
\label{table:exp-robot1}
\end{table}

\section{Experiments} \label{sec:exps}

\subsection{Same-Instance Matching Results} \label{sec:apc-results}

We run all baselines and variants of our method across the APC same-instance matching settings (Section \ref{sec:apc-exp}), with results given in the left side of Table \ref{table:exp-main}. We report accuracy as the total correct matches over the total number of possible correct matches, which we take to be the intersection of the labels of objects visible in any two considered scenes. 

We first note that all proposed matching models do indeed experience performance degradation as difficulty is increased. We further note the surprisingly competitive performance of colour histograms on the same-instance problem, suggesting the relatively simple visual nature of current robotics rearrangement challenges. 

We observe that the \textbf{ResNet50-S} spatial features perform best on the \textbf{Easy} and \textbf{Medium} APC settings. In these cases, with relatively similar source and target views, spatial correspondences between features are useful. For the \textbf{Hard} and \textbf{Hardest} settings, the \textbf{ResNet50-G} global features, which contain no spatial information, perform better. We further note that, across the same-instance matching settings, ResNet50 features trained on ImageNet beat \textbf{CLIP-SemFeat} models. Intuition for this can be gained by analysing the results of \textbf{CLIPVisual}, which uses only the visual embedding of the CLIP model for matching, and which also consistently under-performs ResNet50 ImageNet features. This is likely because ImageNet models are trained with strong data augmentation, and thus are explicitly trained to learn features which are invariant to the kind of visual shifts observed in the same-instance matching setting. In contrast, CLIP is trained only with random-crop data augmentation, with invariances in the visual embedding learned only implicitly through the scale of the training data.

However, even in the same-instance setting, giving the CLIP model access to semantics boosts its performance except in the most trivial setting, and performance degradation as difficulty is increased is much diminished compared to for purely visual approaches. When looking at the \textbf{APC-Easy} setting, \textbf{CLIP-SemFeat} models under-perform \textbf{ResNet50-G} by as much as 16\%, while this is gap is reduced to just 3\% for the \textbf{APC-Hardest} setting.  

Finally, we note that \textbf{CLIP-SemFeat-K}, which uses prompts relating to all 39 objects in the APC dataset for matching, outperforms \textbf{CLIP-SemFeat-N}, which uses only the prompts relating to the objects known to be present in both scenes. One possible explanation for this is found in the interpretation of the visual-semantic matching process (shown in Figure \ref{fig:semfeat}) as employing the projection of visual features onto \textit{directions} given by semantic embeddings. While the directions given by semantic embeddings of objects actually \textit{present} are likely to be by far the most discriminative for matching, the extra directions in \textbf{CLIP-SemFeat-K} can be thought of as providing arbitrary additional scalar projections of the visual features, which leads to extra dimensions in the visual-semantic similarity matrices $C_{s}$, $C_{t}$. We conjecture that these additional dimensions are beneficial to matching in cases where CLIP is not able to relate an object crop to its semantic prompt (i.e. for examples where classification would be inaccurate), such as when object crops are small or highly occluded, as happens often in the APC dataset.

\subsection{Text Prompt Engineering} \label{sec:prompteng}
One appealing feature of the APC dataset is that the objects present are described by their product catalogue names. This presents a challenging test of the extent to which vision-language models such as CLIP can interpret semantic descriptions that are highly specific, and not designed explicitly for downstream applications such as classification or semantically grounded matching. For illustration, 32 of the 39 labels include brand names that carry little intrinsic semantic value. 
Using these labels directly allows us to investigate whether such un-tailored semantic text prompts have a detrimental effect on classification performance, and by extension visual-semantic matching. For comparison, we run K-way classification again using a different set of text prompts, where for each object class we have crudely found an improved text description (see Section \ref{sec:methods}). The classification results from leveraging these in place of the original APC product names is shown in Table \ref{table:classification} as \textbf{CLIP+}. Finally we examine the effect of ensembling over multiple text prompts for each class, which has been shown to be beneficial for CLIP-based classification \cite{Radford2021}. We use all of the short descriptions written in the process of producing \textbf{CLIP+}, and format them into the prompts ``\textit{A picture of a} \{...\}", ``\textit{A picture of a} \{...\}, \textit{a product}", ``\textit{A} \{...\}, \textit{a product}", ``\{...\}". This results in around 20 text prompts per class for ensembling. Results are in Table \ref{table:classification} as \textbf{CLIP++}. We find that \textbf{CLIP+} outperforms \textbf{CLIP} by $7.3$\% in top-1 accuracy, while we see no boost from the further ensembling. The marked performance boost came from a few minutes of trialling prompts for each object, and affirms the significance of the choice of semantic description used in CLIP zero-shot classification, and by extension our visual-semantic matching approaches. We use the prompts arrived at for \textbf{CLIP+} in our APC dataset matching experiments.

\begin{figure*}[ht!]
\includegraphics[width=\textwidth]{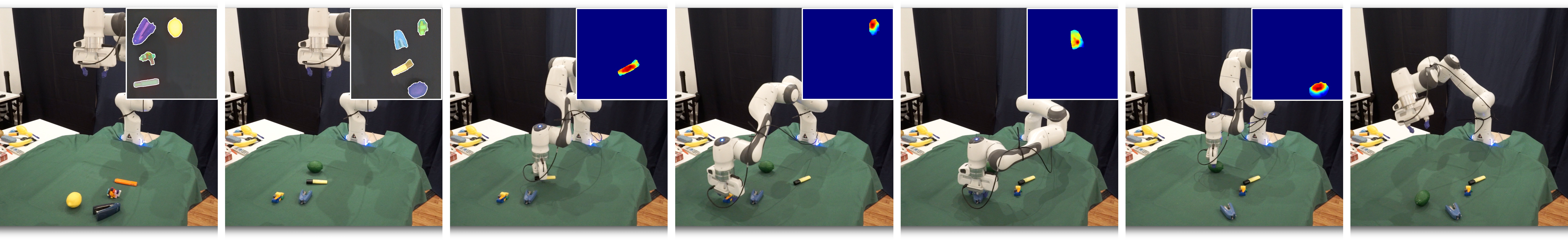}
  \vspace{-1.5em}
  \caption{Successful completion of a different-instance rearrangement task using \textbf{CLIP-SemFeat-N} with labels \texttt{\{highlighter, stapler, lego blocks, citrus fruit\}}. \textit{Inset}: segmentation masks (leftmost frames) and masked GG-CNN grasp predictions. A video of this task is available at \url{https://youtu.be/9soVArIXJlM}.}
  \vspace{-1.5em}
  \label{fig:rollout}
\end{figure*}

\subsection{Cross-Instance Matching Results} \label{sec:lvis-results}
We run all baselines and variants of our method across 20,000 matching scenarios drawn from the LVIS setting described in Section \ref{sec:lvis-exp}, and present results on the right of Table \ref{sec:lvis-exp}.
For text prompts for our visual-semantic matching methods, we use LVIS' object class names (mean length of 1.3 words), formatted as ``\textit{A picture of a} \{...\}". As our LVIS matching setting affords us the possibility of constructing matching problems with up to 40 different object classes at once, we run with $N=8$, which is in line with the higher end of APC matching problems, and $N=20$, to investigate the effect of a fundamentally harder matching setting on matching performance. Average accuracies across methods are reported in Table \ref{table:exp-main}. \textbf{CLIP-SemFeat-K} uses all 40 object class names, while \textbf{CLIP-SemFeat-N} uses only those of the objects present. 

We first note that \textit{all} methods relying purely on visual descriptors perform worse in this setting when compared to the \textbf{APC-Hardest} results, while the \textbf{CLIP-SemFeat} methods achieve the highest accuracy on these cross-instance matching problems. These results confirm that purely visual descriptors struggle to perform good matching across different object instances, reinforcing the importance of leveraging semantics in this setting. It is notable that the accuracy of \textbf{CLIP-SemFeat} methods \textit{improves} between   \textbf{APC-Hardest} and the 8-way cross-instance experiments, despite both the challenging object mismatch and a larger number of objects, on average, to match. This is best explained by the clearer nature of the object crops from LVIS, which are generally less occluded, and also by the succinctness of the semantic labels for LVIS objects, which are more likely to be within the CLIP training distribution than the more verbose and specific product names used in APC settings. Relatedly, and in contrast to the same-instance results, here the \textbf{CLIP-SemFeat-N} method clearly outperforms \textbf{CLIP-SemFeat-K}. Following the discussion in Section \ref{sec:apc-results}, an explanation is that the combination of simpler semantic labels and clearer visual inputs are more likely to be within the CLIP training distribution, and thus the directions provided by the $N$ semantic embeddings of objects to be matched are likely to be highly discriminative, with additional `arbitrary' directions used in the $K$-label setting both redundant and actively detrimental for matching.
\begin{figure}[tbp]
\centering
  \includegraphics[width=0.90\columnwidth]{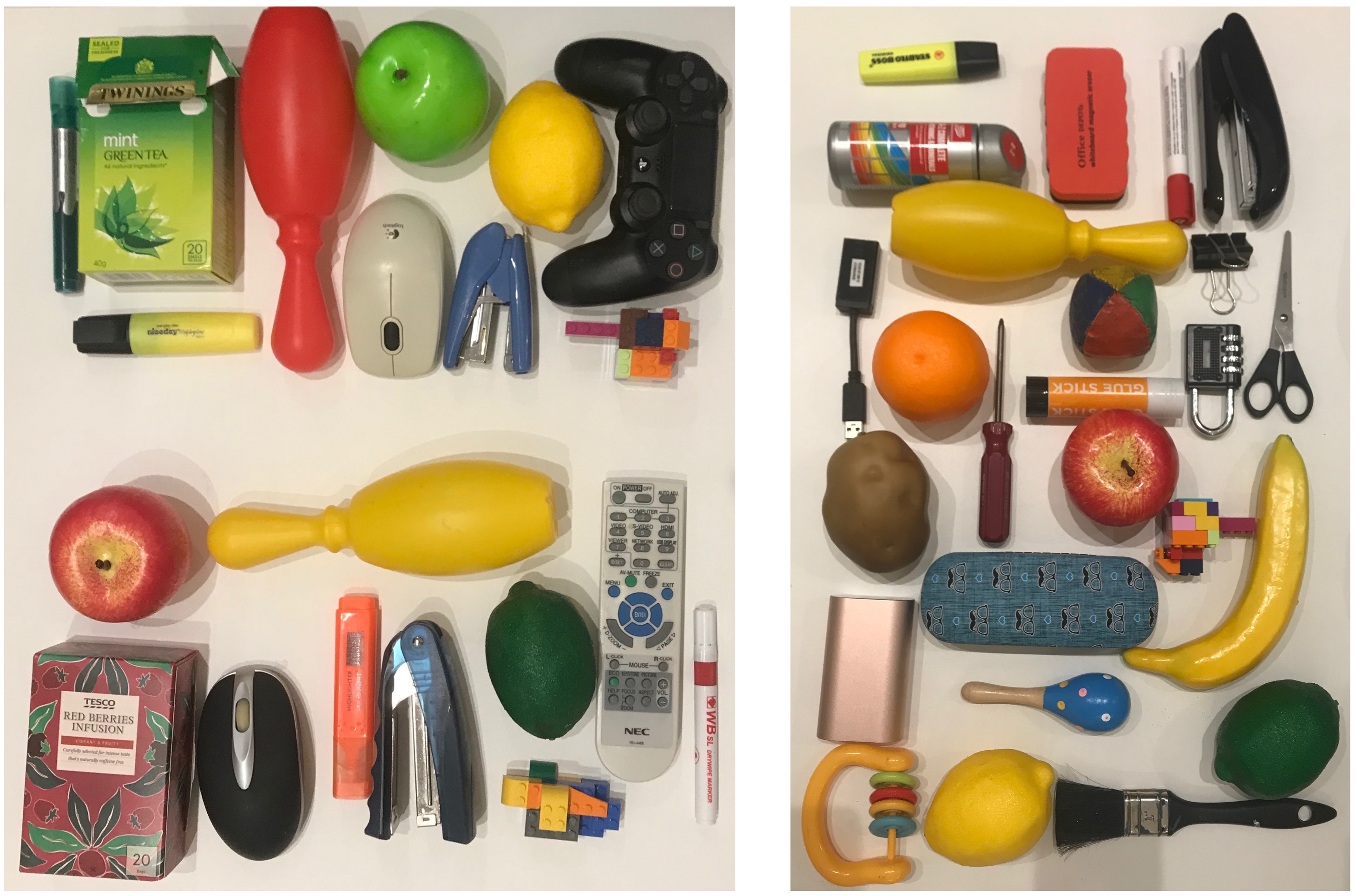}
  \caption{Objects used for robotic rearrangement experiments. (Left) `Twinned' objects, with two visually distinct examples of 10 object classes, used in cross-instance matching experiments. (Right) 25 household objects used for same-instance matching experiments.}
  \label{fig:all-objects}
\end{figure}

\begin{figure}[tbp]
\centering
  \includegraphics[width=0.98\columnwidth]{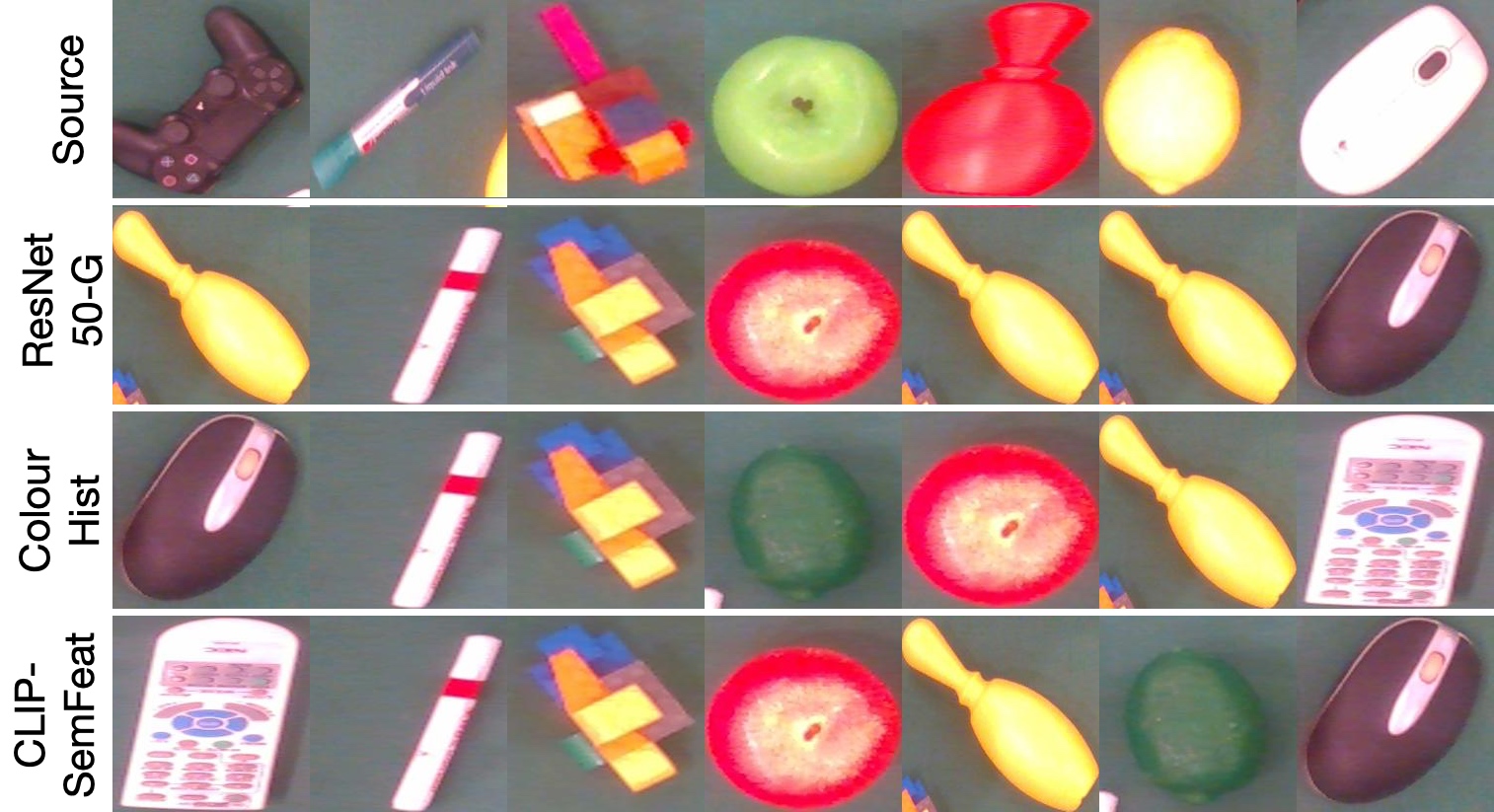}
  \caption{Matching results in the real robot setting for three different matching models. Colour histogram matching tends to conflate similarly coloured objects, while \textbf{ResNet50-G} features exhibit mode collapse, with one target object matched to many source objects. \textbf{CLIP-SemFeat} is able to match all crops accurately.}
  \label{failure_modes}
\vspace{-1.5em}
\end{figure}

\subsection{Real Robot Deployment} \label{sec:robot-exp}
We assess the practical impact of visual-semantic matching across a number of multi-object, multi-step tabletop object rearrangement tasks with a Franka Emika Panda robot arm. As before, we consider both the same-instance and cross-instance matching setting, and are able to address both. Critically, we show that in the latter case we are able to conduct robotic object rearrangement that satisfies a goal image in which all object instances are substantially visually distinct to those in front of the robot, through the use of our semantically grounded matching method.

We collect a set of 25 household objects. We then sample 20 scenes of objects at random, with between 2 and 10 objects per scene. For each object set, we throw the objects into the robot's workspace, with semi-random placement: while we seek occasional occlusions, handling dense clutter is not the focus of this work. An RGB-D image is taken with a robot-wrist-mounted Intel RealSense D435i camera, and segmented with an instance segmentation network \cite{XiangUCN2020}. This is taken to be the goal image. We then conduct 3 further rearrangements of the scene to simulate different starting conditions, taking RGB-D images and producing segmentation masks of each. This gives us 60 source-target image pairs with 395 valid object matches. 

We produce a 2nd set of `twinned' objects for assessing cross-instance matching, consisting of 10 \textit{pairs} of household objects that are different instances of the same item. We produce 20 image pair problems with between 2 and 10 objects per scene. Matching results for both settings are in Table \ref{table:exp-robot1}. \textsc{CLIPdiscrete} assigns labels to objects in the images independently, and takes same-labelled crops as matched. We use the GG-CNN for grasping \cite{Morrison2020}, and write a mask-based collision planner for planning pick-and-place sequences. An example of a successful rearrangement task is shown in Figure \ref{fig:rollout}.

\subsection{Qualitative Matching Model Comparisons}
Figure \ref{failure_modes} provides a typical example of matching in the real-robot setting. Matching results are shown from a single set of source objects to a set of target objects, for three different matching models. Looking first at the third row, it is (unsurprisingly) evident that colour histograms quickly fail once the colour of the source and target objects are dissimilar. For instance, the green apple is matched to the green citrus, while the yellow citrus is matched to the yellow bowling pin. The \textbf{ResNet50-G} features overcome this somewhat, matching mice and apples of different colours, exhibiting some semantic understanding (notably, both these classes feature in the ImageNet training data). However, the \textbf{ResNet50-G features} are not reliably discriminative, and assign the yellow bowling pin in the target image to multiple source image crops. Finally, the CLIP model is most consistently able to leverage semantic information in source and target crops, assigning correct matches despite significantly different visual appearance (for instance between buttoned controllers and differently coloured mice). 

\section{CONCLUSIONS} \label{sec:conclusions}
We examine the problem of object matching for robotic rearrangement tasks instructed with goal images, and characterise the deterioration of existing matching approaches as domain shift between source and target scene increases. We propose a novel approach to matching that makes use of semantic grounding, via a large pre-trained vision-language model, providing additional information about object similarity between scenes. We demonstrate, on both a large-scale dataset and a set of objects in the lab, that this approach enables successful matching even when the objects in source and target scenes are different instances. We integrate  our approach as part of a robotic tabletop object rearrangement system, and were able to complete rearrangement tasks with these cross-domain goal images. We believe that our results motivate further exploration of semantics as a disambiguating factor in vision-based robotic manipulation tasks. 
\section*{ACKNOWLEDGMENT}
The authors gratefully acknowledge the use of the University of Oxford Advanced Research Computing (ARC) facility in carrying out this work (\url{http://dx.doi.org/10.5281/zenodo.22558}).











\bibliographystyle{IEEEtran}
\bibliography{matching.bib}

\end{document}